\documentclass[11pt,a4paper]{article}
\usepackage[hyperref]{emnlp-ijcnlp-2019}
\usepackage{graphicx}
\usepackage{subcaption}
\usepackage{times}
\usepackage{latexsym}
\usepackage{amsmath}
\usepackage{placeins}
\usepackage{url}
\usepackage{multirow}

\aclfinalcopy 

\title{A Closer Look At Feature Space Data Augmentation For Few-Shot \\ Intent Classification}

\author{Varun Kumar, Hadrien Glaude, Cyprien de Lichy,  William Campbell  \\
  Amazon Alexa \\
  Cambridge, MA, USA \\
  {\tt \{kuvrun,hglaude,cllichy,cmpw\}@amazon.com} \\}

\date{}

\begin{document}
\maketitle

\begin{abstract} 
New conversation topics and functionalities are constantly being added to conversational AI agents like Amazon Alexa and Apple Siri. As data collection and annotation is not scalable and is often costly, only a handful of examples for the new functionalities are available, which results in poor generalization performance. We formulate it as a Few-Shot Integration (FSI) problem where a few examples are used to introduce a new intent. In this paper, we study six feature space data augmentation methods to improve classification performance in FSI setting in combination with both supervised and unsupervised representation learning methods such as BERT. Through realistic experiments on two public conversational datasets, SNIPS, and the Facebook Dialog corpus, we show that data augmentation in feature space provides an effective way to improve intent classification performance in few-shot setting beyond traditional transfer learning approaches. In particular, we show that (a) upsampling in latent space is a competitive baseline for feature space augmentation (b) adding the difference between two examples to a new example is a simple yet effective data augmentation method. 
\end{abstract}

\section{Introduction}
Virtual artificial assistants with natural language understanding (NLU) support a variety of functionalities. Throughout the lifespan of the deployed NLU systems, new functionalities with new categories, are regularly introduced. While techniques such as active learning~\cite{peshterliev2018active}, semi-supervised learning~\cite{chodiversity} are used to improve the performance of existing functionalities, performance for new functionalities suffers from the data scarcity problem. 

Recently, Few-Shot Learning has been explored to address the problem of generalizing from a few examples per category. While it has been extensively studied \cite{koch2015siamese,proto,matching} for image recognition, a little attention has been paid to improve NLU performance in the low-data regime. Moreover, researchers have been mostly working on the unrealistic setting that considers tasks with few categories unseen during (pre)training, each with only a few examples, and introduces new categories during test time. We argue that a more realistic setting is Few-Shot Integration (FSI) where new categories with limited training data are introduced into an existing system with mature categories. FSI is well aligned with the goal of lifelong learning of conversational agents and measures the performance in a real-life system setting when only a few examples of a new class are added to the existing data from the old classes.  
To address the poor generalization in data scare scenarios, several pre-training methods such as ELMo~\cite{peters2018deep}, Generative pre-trained Transformer~\cite{radford2018improving}, BERT ~\cite{devlin2018bert}, have been proposed which are trained on a large amount of unannotated text data. Such pre-trained models can be fine-tuned on a particular NLP task and have shown to greatly improve generalization. However, in FSI setting where only a handful of examples are provided, building accurate NLU model is still a challenging task.

In this paper, we focus on Feature space Data Augmentation (FDA) methods to improve the classification performance of the categories with limited data. We study six widely different feature space data augmentation methods: 1) upsampling in the feature space \textsc{Upsample}, 2) random perturbation \textsc{Perturb}, 3) extrapolation~\cite{Devries2017DatasetAI} \textsc{Extra}, 4) conditional variational auto-encoder (CVAE)~\cite{kingma2013auto} \textsc{CVAE}, 5) delta encoder that have been especially designed to work in the few-shot learning setting~\cite{deltaenc2018} \textsc{Delta}, 6) linear delta which is a linear version of the delta encoder \textsc{Linear}. 
While \textsc{Upsample}, \textsc{Perturb}, \textsc{Extra} and \textsc{Linear} doesn't require any training beyond hyper-parameter optimization, \textsc{Delta} and \textsc{CVAE} are trained deep neural network generators.

We compare these six FDA techniques on two open datasets for Intent Classification (IC) : SNIPS \cite{coucke2018snips} and Facebook Dialog corpus \cite{gupta2018semantic}. We show that BERT combined with \textsc{Linear} data augmentation provides an effective method to bootstrap accurate intent classifiers with limited training data.  We make the following contributions:

\begin{enumerate}
    \item We propose the FSI evaluation, a relaxation of the few-shot learning setting that aims to better model the requirement of modern NLU systems. We provide a comprehensive evaluation of FSI for text classification and show that \textsc{Upsample} and \textsc{Perturb} are simple yet efficient baselines that are often neglected in few-shot learning evaluations. 
    
    \item We provide an in-depth analysis of various FDA methods. We show that complex methods such as \textsc{Delta} and \textsc{CVAE} do not always improve over simple methods like \textsc{Linear}, and the performance heavily depends on the feature extractor.
    
    \item Finally, we provide guidance on when and how to apply FDA for FSI. We show that FDA consistently provides gains on top of the unsupervised pre-training methods such as BERT in FSI setting.
\end{enumerate}
\section{Related work}
\textbf{Few-shot learning} has been studied extensively in the computer vision domain. In particular, several metric learning based methods \cite{koch2015siamese,matching,proto,rippel2015metric} has been proposed for few-shot classification where a model first learns an embedding space and then a simple metric is used to classify instances of new categories via proximity to the few labeled training examples embedded in the same space. In addition to metric-learning, several meta-learning based approaches ~\cite{ravi2016optimization,li2017meta,finn2017model} have been proposed for few-shot classification on unseen classes.  

Recently, Few-Shot Learning on text data has been explored using metric learning~\cite{yu2018diverse,jiang2018attentive}. In~\cite{yu2018diverse}, authors propose to learn a weighted combination of metrics obtained from meta-training tasks for a newly seen few-shot task. Similarly, in~\cite{cheng2019fewshot}, authors propose to use meta-metric-learning to learn task-specific metric that can handle imbalanced datasets.

\textbf{Generative models} are also widely used to improve classification performance by data augmentation. For example, generative models are used for data augmentation in image classification ~\cite{mehrotra2017generative,antoniou2018data,metagan2018}, text classification~\cite{gupta2019data}, anomaly detection \cite{lim2018doping}. Data augmentation through deformation of an image has been known to be very effective for image recognition. More advanced approaches rely on Auto-Encoders (AEs) or Generative Adversarial Networks (GANs). For example, in~\cite{mehrotra2017generative} the authors combine metric-learning with data augmentation using GANs for few-shot learning. However, classical generative approaches require a significant amount of training data to be able to generate good enough examples that will improve classification accuracy. To overcome this challenge, \citep{hariharan2017lowshot} proposed to augment the training data in the feature space. This both eases the generation problem and enforces generation of discriminative examples. In addition, the authors propose to transfer deformations from base classes to new classes, which allows circumventing the data scarcity problem for new classes. Finally, in ~\cite{deltaenc2018}, authors used an Autoencoder to encode transformations between pairs of examples of the same class and apply them to an example of the new class.

Generative models are a good candidate for FSI tasks, as one can just combine the generated data for new classes with the old classes training data ~\cite{hariharan2017lowshot,wang2018low}. For text classification, several text generation based data augmentation techniques have also been explored ~\cite{hou2018sequence,zhao2019data,guu2018generating,yoo2018data,cho2019paraphrase}. However, generating discrete sequences, e.g. text, is known to be quite difficult and requires lots of training data. That is why, in this paper, we focus on generative models, which augment data in latent(feature) space to solve a few-shot integration problem for text classification.
\section{Data Augmentation in Feature Space}
\begin{figure}
    \centering
    \begin{subfigure}{1.0\linewidth}
        \centering
        \includegraphics[width=\linewidth]{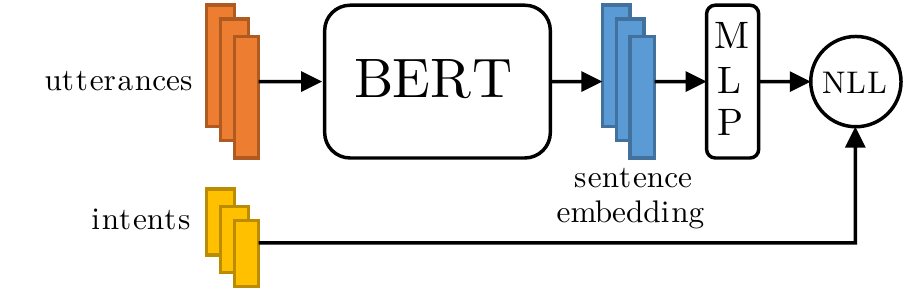}
        \caption{Learning the feature space}
        \label{fig:bert_feature}
    \end{subfigure}
    \begin{subfigure}{1.0\linewidth}
        \centering
        \includegraphics[width=\linewidth]{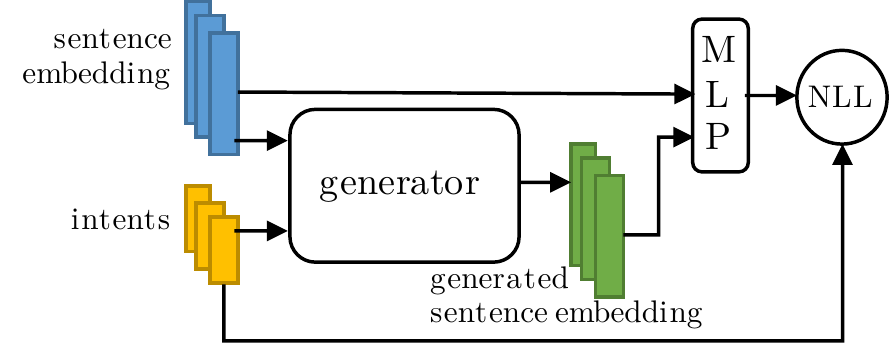}
        \caption{Learning with augmented data}
        \label{fig:augmented_classifier}
    \end{subfigure}
    \caption{Data augmentation in the feature space}
        \label{fig:fda}
\end{figure}

Feature space data Augmentation (FDA) is an effective method to improve classification performance on different ML tasks~\cite{chawla2002smote,hariharan2017lowshot,Devries2017DatasetAI,guo2019augmenting}. As shown on Figure~\ref{fig:fda}, FDA techniques usually work by first learning a data representation or feature extractor, and then generating new data for the low resource class in the feature space. After generating data, a classifier is trained with real and augmented data.

For IC, we finetune a pre-trained English BERT-Base uncased model \footnote{https://github.com/huggingface/pytorch-transformers} to build our feature extractor. The BERT model has $12$ layers, $768$ hidden states, and $12$ heads. We use the pooled representation of the hidden state of the first special token ([CLS]) as the sentence representation. A dropout probability of $0.1$ is applied to the sentence representation before passing it to the 1-layer Softmax classifier. BERT Encoder and MLP classifier are fine-tuned using cross-entropy loss for IC task. Adam~\cite{kingma2014adam} is used for optimization with an initial learning rate of $5e-5$. 

For data augmentation, we apply six different FDA methods, described below, to generate new examples in the feature space. Finally, we train a 1- layer Softmax classifier as in the feature learning phase. 

\subsection{Upsampling}
The simplest method to augment training data for underrepresented categories is to duplicate the existing training data. Upsampling is a well studied technique to handle the class imbalance problem \cite{estabrooks2004multiple}. We show that for intents with limited labeled data, upsampling the existing data in latent space consistently improves model performance, and thus is a good baseline method for FDA techniques. We call this method \textsc{Upsample}.

\subsection{Random Perturbation}
Adding random noise to the existing training data is another simple yet effective data augmentation technique. Random perturbation data augmentation has been previously used to improve the performance of classification models as well as for sequence generation models. For example,~\cite{kurata2016labeled} applied additive and multiplicative perturbation to improve the text generation for data augmentation. In our experiments, we apply both additive and multiplicative perturbation to the existing training data. We sample noise from a uniform distribution [-1.0, 1.0]. We use \textsc{Perturb} to refer to this method. 

\subsection{Conditional VAE} 
Conditional Variational Autoencoder (CVAE) is an extension of Variational Autoencoder (VAE)~\cite{kingma2013auto} which can be used to generate examples for a given category. All components of the model are conditioned on the category.  First, we train CVAE on the sentence representations and then generate new examples by sampling from the latent distribution. The encoder and decoder sub-networks are implemented as multi-layer perceptrons with a single hidden layer of $2048$ units, where each layer is followed by a hyperbolic tangent activation. The encoder output $Z$ is $128$-dimensional.  Mean Square Error (MSE) loss function is used for reconstruction. All models are trained with Adam optimizer with the learning rate set to $10-3$.

\subsection{Linear Delta}
A simple method to generate new examples is to first learn the difference between a pair of examples, and then add this difference to another example. In this case, we first compute the difference $X_i - X_j$ between two examples from the same class and then add it to a third example $X_k$ also from the same class as shown in~\eqref{eqn:linear}. We use \textsc{Linear} to refer to this method. 

\begin{equation}
     \hat{X} = (X_i - X_j) + X_k
     \label{eqn:linear}
\end{equation}

\subsection{Extrapolation}
In~\cite{Devries2017DatasetAI}, authors proposed to use extrapolation to synthesize new examples for a given class. They demonstrated that extrapolating between samples in feature space can be used to augment datasets. In extrapolation, a new example, $\hat{X}$ is generated according to~\eqref{eqn:extra}. In our experiments, we use $\lambda = 0.5$. We call this method \textsc{Extra}.
\begin{equation}
     \hat{X} = (X_i - X_j)* \lambda + X_i
     \label{eqn:extra}
\end{equation}

\subsection{Delta-Encoder}
Delta-Encoder~\cite{deltaenc2018} extends the idea of learning differences between two examples using an autoencoder-based model. It first extracts transferable intra-class deformations (deltas) between same-class pairs of training examples, then applies them to a few examples of a new class to synthesize samples from that class. Authors show that Delta-Encoder can learn transferable deformations from different source classes which can be used to generate examples for unseen classes. While the authors used Delta-Encoder to generate examples for unseen classes, in our experiments, for FSI, we also use the examples from the target class to the train both the feature extractor and the Delta-Encoder along with all other examples. Then we generate new examples for the target category using trained delta encoder. For data generation, we try two different approaches to select a source sentence pair. 
\begin{enumerate}
    \item \textbf{DeltaR}: Sample a pair of sentences ($X_i$, $X_j$) from a randomly selected class. \textsc{DeltaR} applies deltas from multiple source categories to synthesize new examples.
    \item \textbf{DeltaS}: Sample a pair of sentences ($X_i$, $X_j$) from the target category. \textsc{DeltaS} applies deltas from the same target category.  
\end{enumerate}

The encoder and decoder sub-networks are implemented as multi-layer perceptrons with a single hidden layer of $512$ units, where each layer is followed by a leaky ReLU activation ($max(x, 0.2*x)$). The encoder output $Z$ is $16$-dimensional. L1 loss is used as reconstruction loss. Adam optimizer is used with a learning rate of $10-3$. A high dropout with a $50$\% rate is applied to all layers, to avoid the model memorizing examples.

\section{Experiment}
\subsection{Datasets}
We evaluate different FDA techniques on two public benchmark datasets, SNIPS \cite{coucke2018snips}, and Facebook Dialog corpus (FBDialog) \cite{gupta2018semantic}. For SNIPS dataset, we use train, dev and test split provided by \cite{goo2018slot} \footnote{https://github.com/MiuLab/SlotGated-SLU}.

SNIPS dataset contains $7$ intents which are collected from the Snips personal voice assistant. The training, development and test sets contain $13,084$, $700$ and $700$ utterances, respectively. FBDialog has utterances that are focused on navigation, events, and navigation to events. FBDialog dataset also contains utterances with multiple intents as the root node. For our experiment, we exclude such utterances by removing utterances with \textit{COMBINED intent} root node. This leads to $31,218$ training, $4,455$ development and $9,019$ testset utterances. Note that while SNIPS is a balanced dataset, FBDialog dataset is highly imbalanced with a maximum $8,860$ and a minimum of $4$ training examples per intent. 

\subsection{Simulating Few-Shot Integration}
In virtual assistants, often a new intent development starts with very limited training data. To simulate the integration of a new intent, we randomly sample $k$ seed training examples from the new intent, referred to as target intent, and keep all the data from other intents. We also remove the target intent data from the development set. We train the feature extractor on the resulting training data, and then generate $100$, $512$ examples using different augmentation methods for the target intent. To account for random fluctuations in the results, we repeat this process $10$ times for a given target intent and report the average accuracy with the standard deviation. In all experiments, models are evaluated on the full test set. 

\begin{table}
\centering
\resizebox{0.85\columnwidth}{!}{
\begin{tabular}{llll}
\hline
Size & Method & SNIPS & FBDialog \\
\hline
\multicolumn{2}{l}{No Augmentation} & 98.14 (0.42) & 94.99 (0.18)\\ 
\hline
\multirow{7}{*}{5\%}
& \textsc{Upsample}   &  98.14 (0.47)    & 95.01 (0.16)   \\
& \textsc{Perturb}    &  \textbf{98.26} (0.40)    & 94.98 (0.19)    \\
& \textsc{Linear}     &  98.14 (0.45)    & \textbf{95.02} (0.21)    \\
& \textsc{Extra}      &  98.14 (0.45)    & \textbf{95.02} (0.20)   \\
& \textsc{CVAE}       &  98.14 (0.45)    & 94.98 (0.24)   \\
& \textsc{DeltaR}     &  98.23 (0.46)    & 95.00 (0.22)   \\
& \textsc{DeltaS}     &  \textbf{98.26} (0.42)    & 95.00 (0.20)    \\

\hline
\multirow{7}{*}{10\%}
& \textsc{Upsample}   &  98.14 (0.47)    &  94.94 (0.18)  \\
& \textsc{Perturb}    &  98.23 (0.41)    &  94.98 (0.24)   \\
& \textsc{Linear}     &  98.09 (0.50)    &  \textbf{95.02} (0.18)   \\
& \textsc{Extra}      &  98.11 (0.49)    &  95.01 (0.19)   \\
& \textsc{CVAE}       &  98.20 (0.42)    &  94.99 (0.26)   \\
& \textsc{DeltaR}     &  \textbf{98.26} (0.42)    &  94.99 (0.21)   \\
& \textsc{DeltaS}     &  98.23 (0.42)    &  94.97 (0.22)   \\

\hline
\multirow{7}{*}{20\%}
& \textsc{Upsample}   &  98.14 (0.45)    &  95.02 (0.12)  \\
& \textsc{Perturb}    &  98.14 (0.44)    &  94.99 (0.20)   \\
& \textsc{Linear}     &  98.17 (0.43)    &  95.05 (0.23)   \\
& \textsc{Extra}      &  98.14 (0.45)    &  95.07 (0.11)  \\
& \textsc{CVAE}       &  98.11 (0.44)    &  94.98 (0.23)   \\
& \textsc{DeltaR}     &  \textbf{98.26} (0.40)    &  \textbf{95.08} (0.19)   \\
& \textsc{DeltaS}     &  98.20 (0.46)    &  95.04 (0.22)   \\

\hline
\end{tabular}}
\caption{IC accuracy on SNIPS and Facebook dataset with all training data, reported as \textit{mean (SD)}.}
\label{table:full_data_results}
\end{table}

\section{Results and Discussion}

\subsection{FDA For Data-Rich Classification}
For both datasets, we generate $5\%$, $10\%$, and $20\%$ examples using different FDA methods. Then, we train a classifier using both generated as well as real data. Table \ref{table:full_data_results} shows that augmenting data in feature space provides only minor improvements in classification accuracy. In particular, on SNIPS dataset, \textsc{Pertub} and \textsc{DeltaR} improve accuracy from $98.14$ to $98.26$. On FBDialog dataset, DeltaR provides a minor gain, $95.02$ to $95.08$ over upsample baseline. 

\subsection{Impact Of The Number Of Seed Examples}
\begin{figure}
    \centering
    \includegraphics[width=\linewidth]{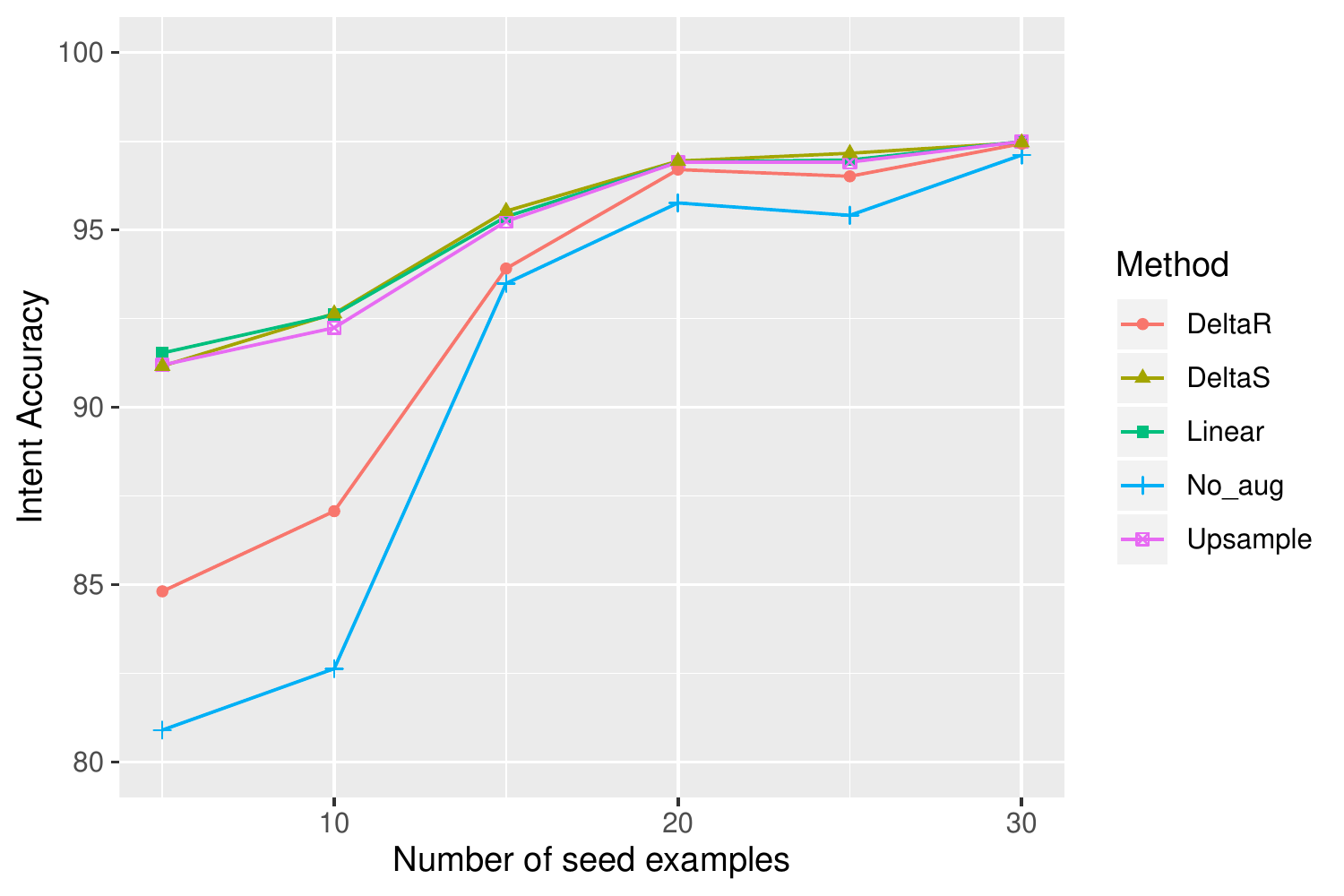}
    \caption{IC accuracy on SNIPS's AddToPlaylist intent with varying number of seed examples. $100$ examples are generated using different FDA techniques. As indicated by the accuracy trend, increasing the seed examples leads to better performance.}
    \label{fig:seed_playlist_100}
\end{figure}

To understand the impact of the number of seed examples, we vary it to $5$, $10$, $15$, $20$, $25$, and $30$ for SNIPS's AddToPlaylist. For each experiment, we generate $100$ examples using different FDA methods. Figure \ref{fig:seed_playlist_100} shows that as the number of seed examples increases, the accuracy of the model goes up. We also observe that for a few seed examples $5$ - $15$, \textsc{Linear} outperforms other FSA methods. Finally, gains are less significant after $30$ seed examples. 

\subsection{Few-Shot Integration} 
We simulate FSI IC for all $7$ intents of SNIPS dataset. For FBDialog dataset, we run simulations on the six largest intents, viz. GetDirections, GetDistance, GetEstimatedArrival, GetEstimatedDuration,  GetInfoTraffic, and GetEvent. Since, BERT generalizes well with just $30$ examples, to compare the effectiveness of different FDA methods, we use $10$ seed examples in FSI simulations. For each intent, we select $k=10$ seed training examples and use all training data for other intents.  

Table~\ref{table:snips_fb_10shot_results} shows average accuracy for all intents' FSI simulations. Results on individual intent's FSI simulations can be found in Appendix's Table~\ref{table:snips_10shot_results} and Table~\ref{table:fb_10shot_results}. On both datasets, all FDA methods improve classification accuracy over no augmentation baseline. Also, \textsc{Upsample} provides huge gains over no augmentation baseline. Additionally, on both datasets, with $512$ augmented examples, \textsc{Linear} and \textsc{DeltaS} works better than \textsc{Perturb} and \textsc{Upsample}. 

\begin{table}
\centering
\resizebox{\columnwidth}{!}{
\begin{tabular}{llll}
 \hline
 \# & Method & SNIPS & FBDialog \\
\hline
\multicolumn{2}{l}{No Augmentation} & 87.46(2.87) & 81.29(0.11)\\ \hline
\multirow{5}{*}{100}
& \textsc{Upsample}  & 94.26(1.66) & \textbf{84.34}(1.84) \\
& \textsc{Perturb}   & 94.18(1.74) & 84.04(1.95)\\
& \textsc{CVAE}      & 94.10(1.83)  & 84.10(1.94)\\
& \textsc{Linear}    & \textbf{94.36}(1.69) & 84.31(1.9)\\
& \textsc{Extra}     & 94.30(1.68) & 84.13(1.83)\\
& \textsc{DeltaR}    & 91.32(3.12) & 81.97(0.76)\\
& \textsc{DeltaS}    & 94.28(1.92)  & 83.50(1.92)\\ \hline

\multirow{5}{*}{512}
& \textsc{Upsample}   & 95.68(0.86) & 89.03(0.99)\\
& \textsc{Perturb}    & 95.65(0.92) & 89.02(0.99)\\
& \textsc{CVAE}       & 95.46(1.03)  & 88.71(1.09)\\
& \textsc{Linear}     & \textbf{95.87}(0.87)  & \textbf{89.30}(1.03)\\
& \textsc{Extra}      & 95.82(0.89)  & 89.21(0.99)\\
& \textsc{DeltaR}     & 95.33(1.56) & 87.28(1.46)\\
& \textsc{DeltaS}     & \textbf{95.88}(1.04)  & 89.15(1.12)\\ \hline

\end{tabular}}
\caption{Average IC accuracy for all intents' FSI simulations on SNIPS and FBDialog dataset. For each simulation, $k=10$ seed examples are used for target intent. Scores are reported as \textit{mean (SD)}. Refer to Appendix's Table~\ref{table:snips_10shot_results} and Table~\ref{table:fb_10shot_results} for individual intents' results.} 
\label{table:snips_fb_10shot_results}
\end{table}
\label{para:few_shot_results}

\subsection{Upsampling: Text Space vs Latent Space}
\begin{table}[t]
\centering
\begin{tabular}{lll}
 \hline
 \# & Method &  Overall Mean  \\
\hline
\multicolumn{2}{l}{No Augmentation} &  94.38(1.23)\\ \hline
\multirow{5}{*}{100}
& \textsc{Upsample}  &  94.53(1.12) \\
& \textsc{Perturb}   & 94.52(1.18) \\
& \textsc{CVAE}     & 94.53(1.18) \\
& \textsc{Linear}    &  94.53(1.12) \\
& \textsc{Extra}     & 94.53(1.13) \\
& \textsc{DeltaR}    & \textbf{94.62}(1.16) \\
& \textsc{DeltaS}    & 94.57(1.14) \\ \hline

\multirow{5}{*}{512}
& \textsc{Upsample}  &  94.67(1.11)\\
& \textsc{Perturb}   &  94.68(1.14) \\
& \textsc{CVAE}      &  94.73(1.11) \\
& \textsc{Linear}    &  94.67(1.11) \\
& \textsc{Extra}     &  94.67(1.11)\\
& \textsc{DeltaR}    &  \textbf{94.88}(1.12)\\
& \textsc{DeltaS}    &  94.74(1.12) \\ \hline
\end{tabular}
\caption{IC accuracy on SNIPS dataset in the FSI setting, reported as \textit{mean (SD)}. The $10$ seed examples are upsampled to $100$ to train the feature extractor. Refer to Appendix's Table~\ref{table:snips_10shot_upsample_results} for  individual intents' results.}
\label{table:snips_10shot_upsample_summary}
\end{table}
In this section, we explore how upsampling in text space impacts performances as it is supposed to both improve the feature extractor and the linear classifier, compared to \textsc{Upsample}. To investigate whether upsampling in text space helps FDA, we upsampled the $10$ seed examples to $100$ and repeat the FSI experiments on all $7$ intents of the SNIPS dataset. Table \ref{table:snips_10shot_upsample_summary} shows the mean accuracy of all $7$ intents FSI simulations results for different FDA techniques. FSI simulations scores for individual intents can be found in Appendix's Table~\ref{table:snips_10shot_upsample_results}. We observe that upsampling in text space improves the no augmentation baseline for all intents. The mean accuracy score improves from $87.46$ to $94.38$. We also observe that different FDA techniques further improve model accuracy. Interestingly, upsampling in text space helps \textsc{DeltaR} the most. Surprisingly, upsampling in latent space provides better performance than upsampling in the text space. In particular, without upsampling the seed examples to learn the feature extractor, the best score is $95.88$ for \textsc{DeltaS}, whereas with text space upsampling the best score decreases to $94.88$. This decrease in performance is only seen with BERT and not with the Bi-LSTM feature extractor (see Table~\ref{table:lstm_results}). We hypothesize that upsampling text data leads to BERT overfitting the target category which results in less generalized sentence representations. Overall, we found that augmentation in the latent space seems to work better with BERT, and is more effective than text space upsampling.  

\subsection{Effect Of The Pre-trained BERT Encoder}
In FSI setting, Fine-Tuned BERT model provides very good generalization performance. For example, for SNIPS's RateBookIntent (column \textit{Book} in Table~\ref{table:snips_10shot_results}), it yields $96.81\%$ accuracy. Overall for BERT representations, \textsc{Linear} and \textsc{DeltaS} augmentation methods provide the best accuracy. 

To investigate whether these augmentation improvements can be generalized to other sentence encoders, we experiment with a Bi-LSTM sentence encoder. For feature learning, we use a 1-layer Bi-LSTM encoder followed by a single layer softmax classifier. In our experiments, we use $128$ as hidden units and $300$ dimension Glove embeddings. For SNIPS dataset, we use $10$ examples of AddToPlaylist intent and for FB Dialog dataset, we use $10$ examples of GetDirections intent. 

Table \ref{table:lstm_results} shows intent accuracy for SNIPS and Facebook datasets. We find that, unlike BERT, in the FSI setting, the Bi-LSTM encoder provides a lower accuracy. In contrast to BERT FSI experiments, \textsc{DeltaS} performs worse than the \textsc{Upsample} and \textsc{Perturb} baselines. The main reason is that Delta-Encoder's performance relies on a good feature extractor and with $10$ seed examples, the Bi-LSTM encoder fails to learn good sentence representations. To improve representation learning, we upsample $10$ utterances to $100$ and then train the feature extractor. Upsampling in text space improves the performance of both delta encoder methods, \textsc{DeltaS}, and \textsc{DeltaR}. Moreover, for both SNIPS's AddToPlayList and FBDialog's GetDirections intent, \textsc{DeltaR} outperforms all other FDA methods. 

\begin{table}
\centering
\resizebox{\columnwidth}{!}{
\begin{tabular}{llllll}
\hline
Size & Method & \multicolumn{2}{l}{SNIPS's AddToPlaylist} &  \multicolumn{2}{l}{FBDialog's GetDirections} \\
\hline
\multicolumn{2}{l}{seed examples ($k$)}& 10  & $100^{*}$ & 10  & $100^{*}$\\
\hline
\multicolumn{2}{l}{No Augmentation} & 80.07 (2.08) & 90.17 (1.39) & 87.44 (0.12) & 87.94 (0.32) \\ 
\hline
\multirow{7}{*}{100}
& \textsc{Upsample}   &  \textbf{88.27} (1.74)    &  90.61 (1.52) & 88.01 (0.26) &  88.17 (0.32)\\
& \textsc{Perturb}    &  88.03 (1.52)    &  90.86 (1.39) & 88.01 (0.32) & 88.25 (0.31)\\
& \textsc{Linear}     &  88.14 (1.62)    &  91.06 (1.58) & 88.05 (0.25) &  88.26 (0.32)\\
& \textsc{Extra}      &  88.09 (1.57)    &  90.74 (1.57) & \textbf{88.10} (0.29) &  88.20 (0.3)\\
& \textsc{CVAE}       &  \textbf{88.27} (2.08)  & 90.90 (1.69) & 88.04 (0.24) & 88.17 (0.32) \\
& \textsc{DeltaR}     &  82.23 (2.21)    &  \textbf{91.46} (1.19) & 87.60 (0.23) & \textbf{88.75} (0.43)\\
& \textsc{DeltaS}     &  84.4 (2.74)    &   91.07 (1.44) & 88.02 (0.22) &  88.57 (0.36)\\
\hline

\multirow{7}{*}{512}
& \textsc{Upsample}   & 91.41 (1.03)    & 91.61 (1.4) & 88.68 (0.49) & 88.40 (0.35)  \\
& \textsc{Perturb}    & \textbf{91.46} (0.99)   & 91.73 (1.32) & 88.89 (0.57) & 88.56 (0.39)\\
& \textsc{Linear}     & 91.20 (1.28)    & 91.41 (1.52) & 88.97 (0.65) & 88.47 (0.33) \\
& \textsc{Extra}      & 91.26 (1.22)    & 91.57 (1.55) & 88.85 (0.61) & 88.48 (0.37)\\
& \textsc{CVAE}       & 91.39 (0.94)    & 91.44 (1.2) & 89.02 (0.52) & 88.48 (0.4)\\
& \textsc{DeltaR}     & 87.09 (2.75)    & \textbf{92.97} (1.2)) & 88.61 (0.35) & \textbf{89.70} (0.53)\\
& \textsc{DeltaS}     & 89.34 (1.48)  & 92.00 (1.25) & \textbf{89.34} (0.4) & 89.09 (0.51)\\
\hline

\end{tabular}}
\caption{IC accuracy on SNIPS's AddToPlaylist and FBDialog's GetDirections in the FSI setting, reported as \textit{mean (SD)}. A $1$-layer Bi-LSTM model is used as a feature extractor. $100^{*}$ represents $10$ seed examples are upsampled to $100$ to train the feature extractor.}
\label{table:lstm_results}
\end{table}

\subsection{Is Delta-Encoder Effective On Text?}
While on few-shot image classification, Delta-Encoder provides excellent generalization performance \cite{deltaenc2018} on unseen classes, on text classification, its performance is heavily dependent on the feature extractor.  We observe that in most cases, \textsc{DeltaR} performs worse than \textsc{DeltaS} which suggests that unlike for few-shot image classification, Delta-Encoder fails to learn variations which can be applied to a different category. In addition, in FSI with BERT encoder, \textsc{DeltaS} performance is close to \textsc{Linear}. This indicates that in the low-data regime, simple subtraction between BERT sentence representations is a good proxy to learn intra-class variations. Upsampling data in text space improves Delta-Encoder performance for both BERT and Bi-LSTM encoders. As shown in Table~\ref{table:snips_10shot_upsample_summary}, with upsampling in text space, \textsc{DeltaR} performs better than any other FDA method. 

\subsection{Qualitative Evaluation}
\begin{figure*}[ht]
    \centering
    \begin{minipage}{0.5\textwidth}
        \centering
        \includegraphics[width=1.0\textwidth, height=8cm]{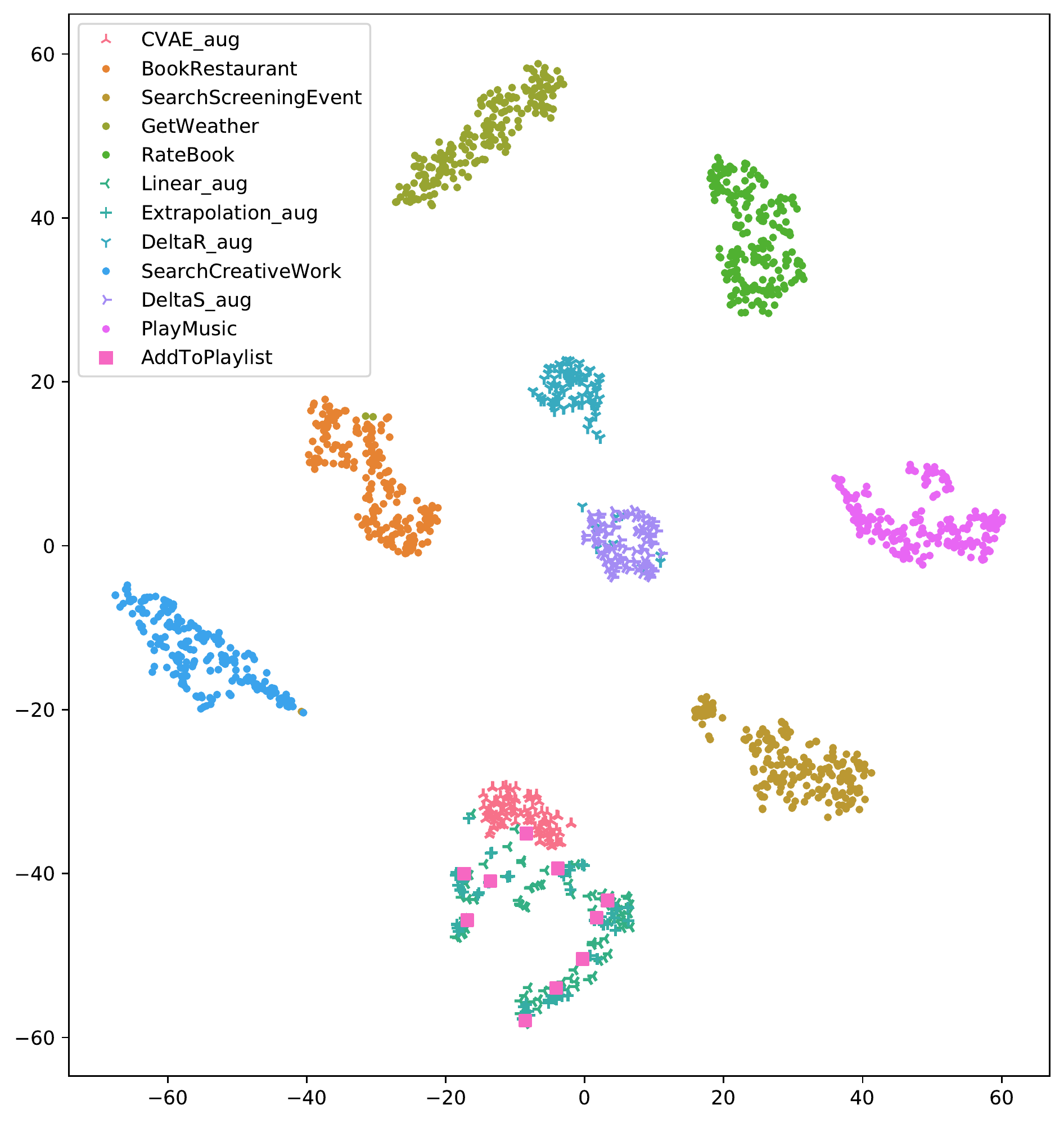}
        \caption{$10$ seed examples}
        \label{fig:addtoplaylist_tsne_10}
    \end{minipage}\hfill
    \begin{minipage}{0.5\textwidth}
        \centering
        \includegraphics[width=1.0\textwidth, height=8cm]{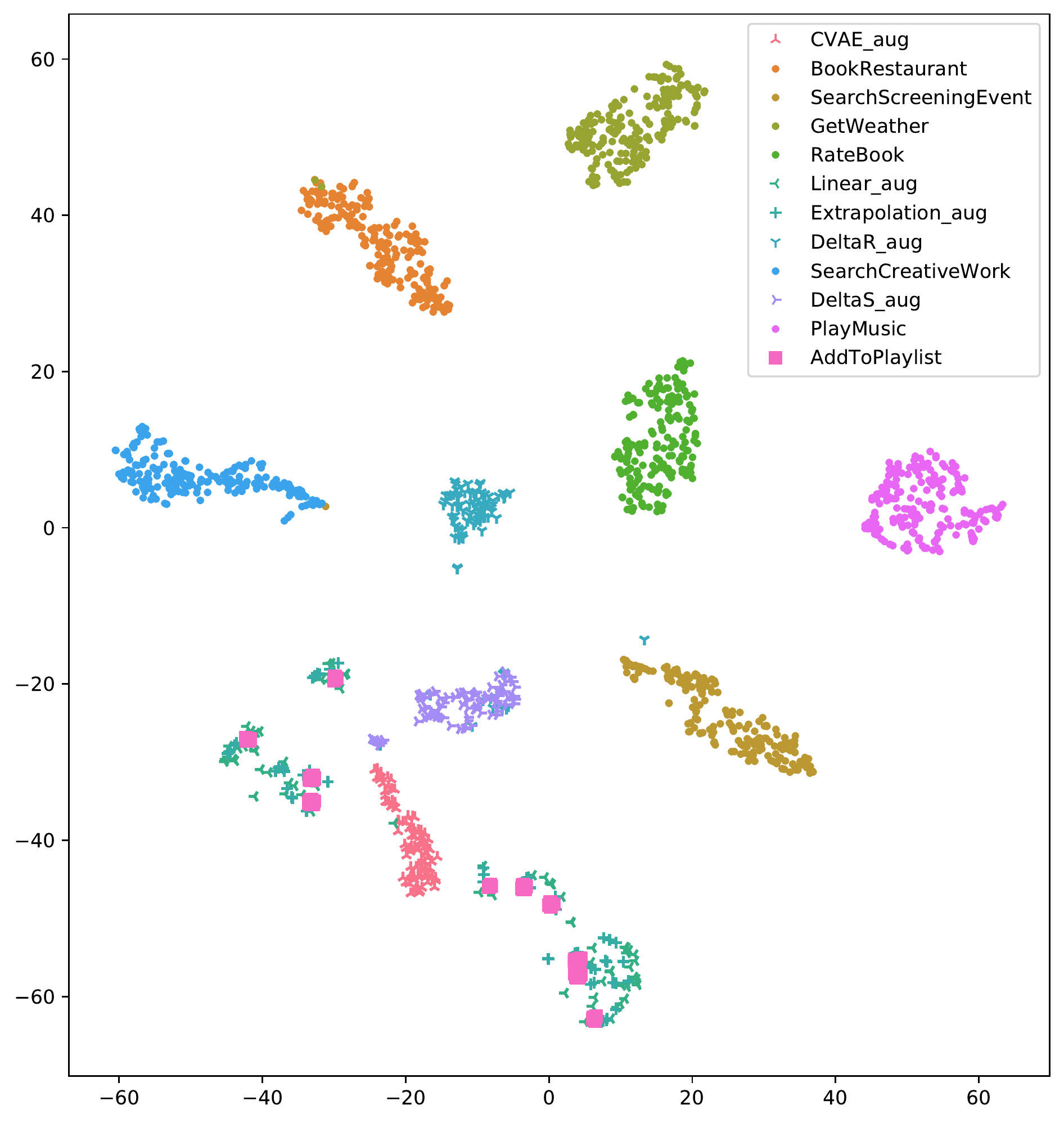}
        \caption{$10$ seed examples are upsampled to $100$}
        \label{fig:addtoplaylist_tsne_10_upsample}
    \end{minipage}
    \caption{t-SNE visualization of different data augmentation methods for AddToPlaylist intent. BERT encoder is used to learn sentence representations.}
        \label{fig:tsne}
\end{figure*}

We observe significant accuracy improvements in all FSI experiments for all FDA methods. Since \textsc{Upsample} and \textsc{Perturb} also provide significant gains, it seems that most of the gains come from the fact that we are adding more data. However, in the FSI setting, \textsc{Linear} and \textsc{DeltaS} method consistently perform better than both \textsc{Upsample} and \textsc{Perturb}, which indicates that these methods generate more relevant data than just noise, and redundancy. Here, we focus on visualizing generated examples from \textsc{Linear}, \textsc{DeltaS} and \textsc{DeltaR} methods using t-SNE. 

Figure \ref{fig:addtoplaylist_tsne_10} shows visualizations for SNIPS's AddToPlaylist generated sentence representations using different FDA methods. We use $10$ seed examples of AddToPlaylist and use BERT as sentence encoder. While data generated by \textsc{Linear} and \textsc{Extra} are close to the real examples, \textsc{DeltaS} and \textsc{DeltaR} generated examples form two different clusters. Since, Delta-Encoder performance improves when seed examples are upsampled in text space, we plot sentence examples from upsampled data. 

Figure \ref{fig:addtoplaylist_tsne_10_upsample} shows that when $10$ seed examples are upsampled to $100$, \textsc{DeltaS} cluster moves closer to the seed examples, and while most of the \textsc{DeltaR} generated data forms a separate cluster, a few of the generated examples are close to the seed examples. Since, in experiments with upsampled text examples, \textsc{DeltaR} performs better than other FDA methods, we hypothesize that \textsc{DeltaR} increases the amount of variability within the dataset by generating diverse examples which leads to a more robust model. 
   

\section{Conclusion and Future Work}
In this paper, we investigate six FDA methods including \textsc{Upsample}, \textsc{Perturb}, \textsc{CVAE}, Delta-Encoder, \textsc{Extra}, and \textsc{Linear} to augment training data. We show that FDA works better when combined with transfer learning and provides an effective way of bootstrapping an intent classifier for new classes. As expected, all FDA methods become less effective when the number of seed examples increases and provides minor gains in the full-data regime. Through comparing methods on two public datasets, our results show that \textsc{Linear} is a competitive baseline for FDA in FSI setting, especially when combined with transfer learning (BERT). 

Additionally, we provide empirical evidence that in few-shot integration setting, feature space augmentation combined with BERT provides better performance than widely used text space upsampling. Given that pre-trained language models provide state of the art performance on several NLP tasks, we find this result to be in particular encouraging, as it shows potential for applying FDA methods to other NLP tasks. 

Our experiments on Delta-Encoder also shows that unlike few-shot image classification, Delta-Encoder fails to learn transferable intra-class variations. This result emphasizes that methods providing improvements in computer vision domain might not produce similar gains on NLP tasks, thus underlining the need to develop data augmentation methods specific to NLP tasks. 

\bibliography{fewshot}
\bibliographystyle{acl_natbib}

\appendix
\section{FSI experiment results for all intents}
In all tables, individual columns represent FSI results for an intent, and \textit{Overall Mean} column, provides average accuracy for all intents' FSI simulations. 

\begin{table*}[t]
\centering
\resizebox{\textwidth}{!}{
\begin{tabular}{llllllllll}
 \hline
 \# & Method & Playlist & Restaurant & Weather & Music & Book  & Work & Event & Overall Mean \\
\hline
\multicolumn{2}{l}{No Augmentation} & 82.63(5.11) & 87.86(3.53) & 84.51(1.3) & 88.07(2.37) & 96.81(2.94) & 85.14(1.53) & 87.19(3.31) & 87.46(2.87)\\ \hline
\multirow{5}{*}{100}
& \textsc{Upsample}  & 92.24(2.96) & 97.7(0.67) & 96.44(0.75) & 94.57(1.1) & 97.96(0.82) & 89.61(3.01) & 91.26(2.35) & 94.26(1.66)\\
& \textsc{Perturb}   & \textbf{93.09}(2.55) & 97.41(0.92) & 96.07(1.35) & 94.39(1.13) & 97.86(0.93) & 89.36(2.76) & 91.09(2.53) & 94.18(1.74)\\
& \textsc{CVAE}      & 92.4(3.66) & 97.47(0.67) & 96.49(1.07) & 94.36(1.26) & 97.71(1.1) & 89.1(2.79) & 91.2(2.22) & 94.1(1.83)\\
& \textsc{Linear}    & 92.61(3.02) & 97.74(0.67) & 96.44(0.77) & 94.63(1.18) & \textbf{97.97}(0.78) & \textbf{89.61}(3.05) & 91.53(2.34) & \textbf{94.36}(1.69)\\
& \textsc{Extra}     & 92.36(3.0) & 97.74(0.66) & 96.41(0.77) & 94.6(1.18) & 97.97(0.78) & 89.47(3.11) & 91.51(2.3) & 94.3(1.68)\\
& \textsc{DeltaR}    & 87.07(4.67) & 93.57(4.07) & 91.0(4.23) & 94.87(1.28) & 97.66(1.42) & 85.97(2.34) & 89.11(3.84) & 91.32(3.12)\\
& \textsc{DeltaS}    & 92.64(4.49) & \textbf{97.76}(0.7) & 96.41(1.25) & \textbf{94.99}(0.92) & 97.83(0.99) & 88.69(2.69) & \textbf{91.64}(2.36) & 94.28(1.92)\\ \hline

\multirow{5}{*}{512}
& \textsc{Upsample}   &  95.3(1.09) & 98.0(0.64) & 97.63(0.34) & 95.57(0.87) & 98.03(0.55) & 92.0(1.49) & 93.26(1.05) & 95.68(0.86)\\
& \textsc{Perturb}    &  95.33(1.2) & 97.94(0.6) & 97.6(0.44) & 95.5(0.91) & 97.91(0.55) & 92.03(1.78) & 93.21(0.99) & 95.65(0.92)\\
& \textsc{CVAE}       & 95.46(1.12) & 97.89(0.62) & 97.54(0.43) & 95.36(1.02) & 97.93(0.7) & 91.34(2.17) & 92.73(1.19) & 95.46(1.03)\\
& \textsc{Linear}     & 95.39(1.1) & \textbf{98.0}(0.64) & 97.67(0.36) & 95.74(0.89) & \textbf{98.04}(0.5) & \textbf{92.61}(1.47) & 93.66(1.13) & \textbf{95.87}(0.87)\\
& \textsc{Extra}      & 95.36(1.17) & 98.0(0.64) & 97.66(0.37) & 95.74(0.88) & 98.04(0.5) & 92.29(1.52) & 93.63(1.17) & 95.82(0.89)\\
& \textsc{DeltaR}     & 95.36(1.74) & 97.81(0.69) & 97.6(0.44) & 95.9(0.97) & 97.74(1.02) & 90.27(3.44) & 92.61(2.64) & 95.33(1.56)\\
& \textsc{DeltaS}     & \textbf{95.66}(1.18) & 97.96(0.59) & \textbf{97.8}(0.45) & \textbf{95.91}(0.88) & 97.91(0.74) & 92.26(2.57) & \textbf{93.66}(0.86) & \textbf{95.88}(1.04)\\ \hline

\end{tabular}}
\caption{IC accuracy on SNIPS dataset in the FSI setting ($k=10$), reported as \textit{mean (SD)}.} 
\label{table:snips_10shot_results}
\end{table*}

\begin{table*}[t]
\centering
\resizebox{\textwidth}{!}{
\begin{tabular}{llllllllll}
 \hline
 \# & Method & Directions &  Distance &  Arrival & Duration &  Traffic & Event & Overall Mean \\
\hline
\multicolumn{2}{l}{No Augmentation} &  89.61(0.1) & 89.94(0.09) & 90.56(0.12) & 81.74(0.13) & 68.5(0.13) & 67.39(0.11) & 81.29(0.11)\\ \hline
\multirow{5}{*}{100}
& \textsc{Upsample}  & 89.89(0.27) & 93.64(0.87) & 92.95(0.57) & 84.28(3.45) & 68.99(0.49) & \textbf{76.26}(5.41) & \textbf{84.34}(1.84) \\
& \textsc{Perturb}   & 89.82(0.24) & 93.58(0.84) & 92.81(0.55) & \textbf{84.81}(3.77) & \textbf{69.15}(0.68) & 74.07(5.6) & 84.04(1.95)\\
& \textsc{CVAE}      & 89.91(0.32) & 93.46(0.77) & 92.7(0.67) & 84.45(3.52) & 69.11(0.9) & 74.94(5.49) & 84.1(1.94)\\
& \textsc{Linear}    & \textbf{89.93}(0.24) & \textbf{93.65}(0.88) & \textbf{92.98}(0.57) & 84.2(3.44) & 68.96(0.51) & 76.12(5.77) & 84.31(1.9)\\
& \textsc{Extra}     & 89.88(0.27) & 93.61(0.89) & 92.96(0.59) & 84.21(3.43) & 68.94(0.46) & 75.18(5.34) & 84.13(1.83) \\
& \textsc{DeltaR}    & 89.64(0.11) & 92.57(1.3) & 90.79(0.37) & 81.72(0.12) & 68.48(0.08) & 68.63(2.59) & 81.97(0.76)\\
& \textsc{DeltaS}    & 89.88(0.34) & 93.68(0.72) & 92.6(0.76) & 83.88(3.2) & 68.93(0.67) & 72.05(5.83) & 83.5(1.92)\\ \hline

\multirow{5}{*}{512}
& \textsc{Upsample}  & 91.93(0.48) & 94.58(0.34) & 93.99(0.31) & 92.56(0.72) & 75.84(2.19) & 85.27(1.87) & 89.03(0.99)\\
& \textsc{Perturb}   & 91.78(0.49) & 94.58(0.43) & 94.02(0.25) & 92.53(0.87) & \textbf{76.0}(2.27) & 85.22(1.61) & 89.02(0.99)\\
& \textsc{CVAE}      & 91.85(0.52) & 94.57(0.39) & 94.0(0.34) & 92.45(0.92) & 74.91(2.73) & 84.5(1.61) & 88.71(1.09) \\
& \textsc{Linear}    & 92.14(0.66) & 94.6(0.35) & 94.05(0.32) & \textbf{92.78}(0.67) & \textbf{76.0}(2.49) & 86.22(1.7) & \textbf{89.3}(1.03)\\
& \textsc{Extra}     & \textbf{92.11}(0.57) & 94.61(0.35) & 94.04(0.29) & 92.72(0.7) & 75.79(2.45) & 85.98(1.58) & 89.21(0.99)\\
& \textsc{DeltaR}    & 90.43(0.55) & 94.54(0.35) & 93.8(0.3) & 86.64(4.38) & 71.68(1.46) & \textbf{86.55}(1.75) & 87.28(1.46)\\
& \textsc{DeltaS}    & 91.83(0.47) & \textbf{94.66}(0.4) & \textbf{94.08}(0.24) & 92.31(1.45) & 75.81(2.1) & 86.23(2.08) & 89.15(1.12)\\ \hline

\end{tabular}}
\caption{IC accuracy on FBDialog dataset in the FSI setting ($k=10$), reported as \textit{mean (SD)}.}
\label{table:fb_10shot_results}
\end{table*}

\begin{table*}[t]
\centering
\resizebox{\textwidth}{!}{
\begin{tabular}{llllllllll}
 \hline
 \# & Method & Playlist & Restaurant & Weather & Music & Book  & Work & Event & Overall Mean  \\
\hline
\multicolumn{2}{l}{No Augmentation} &  96.0(1.69) & 95.39(1.59) & 96.41(1.18) & 93.1(1.38) & 97.79(0.77) & 88.46(1.14) & 93.49(0.87) & 94.38(1.23)\\ \hline
\multirow{5}{*}{100}
& \textsc{Upsample}  & 96.0(1.57) & 95.87(1.26) & \textbf{96.51}(1.04) & 93.19(1.25) & 97.83(0.7) & 88.63(1.21) & 93.7(0.83) & 94.53(1.12) \\
& \textsc{Perturb}   & 96.1(1.64) & 95.7(1.23) & 96.43(1.28) & 93.33(1.1) & 97.8(0.77) & 88.56(1.32) & 93.7(0.9) & 94.52(1.18) \\
& \textsc{CVAE}      & 96.07(1.46) & \textbf{95.91}(1.43) & 96.43(1.31) & 93.2(1.15) & 97.83(0.78) & 88.63(1.28) & 93.66(0.86) & 94.53(1.18) \\
& \textsc{Linear}    & 96.0(1.57) & 95.89(1.26) & \textbf{96.51}(1.04) & 93.19(1.25) & 97.83(0.7) & 88.63(1.21) & 93.7(0.83) & 94.53(1.12) \\
& \textsc{Extra}     & 96.0(1.57) & 95.84(1.3) & \textbf{96.51} (1.04) & 93.19(1.25) & 97.83(0.7) & 88.63(1.21) & 93.7(0.83) & 94.53(1.13) \\
& \textsc{DeltaR}    & 96.09(1.51) & 95.74(1.46) & 96.44(1.29) & \textbf{93.56}(0.95) & \textbf{97.86}(0.75) & \textbf{88.79}(1.25) & \textbf{93.86}(0.93) & \textbf{94.62}(1.16) \\
& \textsc{DeltaS}    & \textbf{96.11}(1.52) & 95.69(1.44) & 96.46(1.29) & 93.44(0.93) & \textbf{97.86}(0.75) & 88.64(1.18) & 93.76(0.89) & 94.57(1.14) \\ \hline

\multirow{5}{*}{512}
& \textsc{Upsample}  & 96.07(1.54) & 96.09(1.2) & 96.6(1.06) & 93.5(1.14) & \textbf{97.87}(0.69) & 88.73(1.23) & 93.8(0.92) & 94.67(1.11)\\
& \textsc{Perturb}   & 96.23(1.6) & 96.17(1.23) & 96.63(1.13) & 93.49(1.03) & 97.84(0.72) & 88.6(1.3) & 93.79(0.98) & 94.68(1.14) \\
& \textsc{CVAE}      & 96.14(1.46) & 96.24(1.18) & 96.63(1.06) & 93.6(1.08) & 97.87(0.75) & 88.76(1.29) & 93.87(0.98) & 94.73(1.11) \\
& \textsc{Linear}    & 96.07(1.54) & 96.11(1.21) & 96.6(1.06) & 93.49(1.13) & \textbf{97.87}(0.69) & 88.76(1.25) & 93.8(0.92) & 94.67(1.11) \\
& \textsc{Extra}     & 96.07(1.54) & 96.13(1.18) & 96.6(1.06) & 93.5(1.14) & \textbf{97.87}(0.69) & 88.73(1.25) & 93.8(0.92) & 94.67(1.11)\\
& \textsc{DeltaR}    & \textbf{96.29}(1.52) & \textbf{96.29}(1.34) & \textbf{96.71}(1.1) & \textbf{93.87}(1.04) & 97.86(0.75) & \textbf{89.11}(1.22) & \textbf{94.03}(0.89) & \textbf{94.88}(1.12)\\
& \textsc{DeltaS}    & 96.19(1.61) & 96.2(1.23) & 96.69(1.07) & 93.61(0.96) & 97.86(0.75) & 88.84(1.28) & 93.83(0.94) & 94.74(1.12) \\ \hline
\end{tabular}}
\caption{IC accuracy on SNIPS dataset in the FSI setting, reported as \textit{mean (SD)}. The $10$ seed examples are upsampled to $100$ to train the feature extractor.}
\label{table:snips_10shot_upsample_results}
\end{table*}

\end{document}